\documentclass[letterpaper, 10 pt, conference]{ieeeconf}  

\IEEEoverridecommandlockouts                              

\usepackage{graphics} 
\usepackage{epsfig} 
\usepackage{times} 
\usepackage{amsmath} 
\usepackage{amssymb}  
\let\proof\relax 
\let\endproof\relax
\usepackage{amsthm}

\usepackage[T1]{fontenc}
\usepackage{soul}
\usepackage{url}
\usepackage[hidelinks]{hyperref}
\usepackage[utf8]{inputenc}
\usepackage[small]{caption}
\usepackage{booktabs}
\usepackage{algorithm}
\usepackage[noend]{algpseudocode}
\usepackage{algorithmicx}
\usepackage[switch]{lineno}
\usepackage{subcaption}
\usepackage{cleveref}
\usepackage{diagbox}  
\usepackage{booktabs} 

\usepackage[dvipsnames]{xcolor}

\definecolor{mycustomred}{RGB}{204, 10, 10}

\algdef{SE}[SWITCH]{Switch}{EndSwitch}[1]{\algorithmicswitch\ #1\ \algorithmicdo}{\algorithmicend\ \algorithmicswitch}%
\algdef{SE}[CASE]{Case}{EndCase}[1]{\algorithmiccase\ #1}{\algorithmicend\ \algorithmiccase}%
\algtext*{EndSwitch}%
\algtext*{EndCase}%
\algrenewcommand\textproc{}

\algnewcommand\algorithmicinput{\textbf{Input:}}
\algnewcommand\INPUT{\item[\algorithmicinput]}
\algnewcommand\algorithmicoutput{\textbf{Output:}}
\algnewcommand\OUTPUT{\item[\algorithmicoutput]}

\algnewcommand\algorithmicswitch{\textbf{switch}}
\algnewcommand\algorithmiccase{\textbf{case}}
\algnewcommand\algorithmicassert{\texttt{assert}}
\algnewcommand\Assert[1]{\State \algorithmicassert(#1)}%
\algdef{SE}[SWITCH]{Switch}{EndSwitch}[1]{\algorithmicswitch\ #1\ \algorithmicdo}{\algorithmicend\ \algorithmicswitch}%
\algdef{SE}[CASE]{Case}{EndCase}[1]{\algorithmiccase\ #1}{\algorithmicend\ \algorithmiccase}%
\algtext*{EndSwitch}%
\algtext*{EndCase}%

\title{\LARGE \bf
RAILGUN: A Unified Convolutional Policy for\\Multi-Agent Path Finding Across Different Environments and Tasks
}

\author{Yimin Tang$^{1*}$, Xiao Xiong$^{2*}$, Jingyi Xi$^2$, Jiaoyang Li$^3$, Erdem Bıyık$^{1}$, Sven Koenig$^{4}$
\thanks{$^{*}$Equal contribution}%
\thanks{$^{1}$Thomas Lord Department of Computer Science, University of Southern California, {\tt\small yimintan@usc.edu, biyik@usc.edu}}%
\thanks{$^{2}$Independent Researcher, {\tt\small xiaoxiong.xx21@gmail.com, flotherxi@gmail.com}}%
\thanks{$^{3}$Carnegie Mellon University, {\tt\small jiaoyanl@andrew.cmu.edu}}%
\thanks{$^{4}$University of California, Irvine, {\tt\small sven.koenig@uci.edu}}%
}%

\begin{document}

\maketitle
\thispagestyle{empty}
\pagestyle{empty}

\begin{abstract}
Multi-Agent Path Finding (MAPF), which focuses on finding collision-free paths for multiple robots, is crucial for applications ranging from aerial swarms to warehouse automation. Solving MAPF is NP-hard so learning-based approaches for MAPF have gained attention, particularly those leveraging deep neural networks. Nonetheless, despite the community's continued efforts, all learning-based MAPF planners still rely on decentralized planning due to variability in the number of agents and map sizes.
We have developed the first centralized learning-based policy for MAPF problem called RAILGUN. RAILGUN is not an agent-based policy but a map-based policy. By leveraging a CNN-based architecture, RAILGUN can generalize across different maps and handle any number of agents. We collect trajectories from rule-based methods to train our model in a supervised way. In experiments, RAILGUN outperforms most baseline methods and demonstrates great zero-shot generalization capabilities on various tasks, maps and agent numbers that were not seen in the training dataset.
\end{abstract}

\section{Introduction}

Multi-Agent Path Finding (MAPF) is an NP-hard problem~\cite{stern2019multi,yu2013structure} which focuses on finding collision-free paths for multiple agents to move from start locations to their goal locations in a known environment while optimizing a specified cost function. This problem could be adapted to many realistic scenarios from aerial swarms to warehouse automation which are multi-billion dollar industries. Many algorithms have been proposed to solve this problem or its variants, such as Conflict-Based Search (CBS)~\cite{sharon2015conflict}, $M^*$~\cite{wagner2011m}, LaCAM~\cite{okumura2023lacam} and MAPF-LNS2~\cite{li2022mapf}.

As neural networks demonstrate their powerful capabilities in various fields of computer science~\cite{krizhevsky2012imagenet,silver2016mastering,achiam2023gpt}, learning-based MAPF solvers have also garnered significant attention~\cite{alkazzi2024comprehensive}. Currently, all learning-based MAPF solvers adopt decentralized approaches, where each agent takes surrounding local information as input, typically represented as a field-of-view (FOV). These decentralized policies determine each agent's action, either simultaneously or sequentially, at the current timestep based on the agent's FOV input.
Many decentralized methods have been proposed, such as PRIMAL~\cite{sartoretti2019primal}, MAPPER~\cite{liu2020mapper}, MAGAT~\cite{li2021message}, SCRIMP~\cite{wang2023scrimp}, and MAPF-GPT~\cite{andreychuk2024mapf}. These methods primarily rely on imitation learning (IL) and reinforcement learning (RL) and often incorporate additional components, such as inter-agent communication, to enhance performance. It is important to note these approaches focus on individual agents and attempt to generate actions based on agent-specific features, which typically do not include global state information. 
Furthermore, as features are based on the agent itself, these approaches inherently allow the number of agents to vary.

On the other hand, centralized approaches bring several benefits, such as the ability to coordinate the movements of multiple agents. However, the literature lacks centralized MAPF algorithms that are \emph{learning-based}, since it is challenging to train a centralized neural network that can handle variability in both the number of agents and map sizes.

We present RAILGUN the first centralized learning-based method for MAPF which generates actions based on maps rather than individual agents. The core idea of RAILGUN is to generate a directed graph in which each node has exactly one outgoing edge at every timestep. This design enables our method to handle any number of agents on the map. Additionally, we use U-Net~\cite{ronneberger2015u} as the model backbone which produces outputs of the same dimensions as the input features. This allows RAILGUN to accommodate maps of varying sizes.
In summary, our contributions are as follows:
\begin{itemize}
\item We propose the first centralized learning-based MAPF algorithm, RAILGUN, which generates actions for map grid cells rather than for individual agents.
\item We design a CNN-based network enabling RAILGUN to handle maps of different sizes.  
\item Through experiments in diverse test settings, we demonstrate that RAILGUN, trained on data from one map type, generalizes effectively to new types of maps and testing scenarios, and outperforms most baseline methods in POGEMA~\cite{skrynnik2024pogemabenchmarkplatformcooperative} benchmark.
\end{itemize}

\section{Problem Definition}

The MAPF problem is defined as follows: Let $I=\{1,2,\cdots,N\}$ denote a set of $N$ agents. $G = (V,E)$ represents an undirected graph, where each vertex $v \in V$ represents a possible location of an agent in the workspace, and each edge $e \in E$ is a unit-cost edge between two vertices that moves an agent from one vertex to the other. In this paper, we focus on 2D grid maps with connections in four directions. Self-loop edges are also allowed, which represent ``wait-in-place'' actions.
Each agent $i\in I$ has a start location $s_i \in V$ and a goal location $g_i \in V$. It also holds that $s_i \neq s_j$ and $g_i \neq g_j$ when $i \neq j$ $\forall i, j \in I$.
Our task is to plan a collision-free path for each agent $i$ from $s_i$ to $g_i$.

Each action of agents, either waiting in place or moving to an adjacent vertex, takes one time unit.
Let $v^i_t \in V$ be the location of agent $i$ at timestep $t$.
Let $\pi_i=[v_0^{i}, v_1^{i}, ..., v_{T^{i}}^{i}]$ denote a path of agent $i$ from its start location $v_0^{i}$ to its target $v_{T^{i}}^{i}$. 
We assume that agents rest at their targets after completing their paths, i.e., $v_t^i = v_{T^i}^i, \forall t > T^i$.
The cost of agent $i$'s path is $T^i$. We refer to the path with the minimum cost as the shortest path.

We consider two types of agent-agent collisions.
The first type is \emph{vertex collision}, where two agents $i$ and $j$ occupy the same vertex at the same timestep. 
The second type is \emph{edge collision}, where two agents move in opposite directions along the same edge simultaneously.
We use $(i, j, t)$ to denote a vertex collision between agents $i$ and $j$ at timestep $t$ or an edge collision between agents $i$ and $j$ at timestep $t$ to $t+1$.
The requirement of being collision-free implies the targets assigned to the agents must be distinct from each other. We use \emph{SoC (flowtime)} $\sum_{i=1}^{N}T^{i}$ as the cost function.

The objective of the MAPF problem is to find a set of paths $\{\pi_i \mid i\in I\}$ for all agents such that, for each agent $i$:
\begin{enumerate}
\item Agent $i$ starts from its start location (i.e., $v_0^{i} = s_i$) and stops at its target location $g_j$ (i.e., $v_{t}^{i} = g_j, \forall t \ge T^{i}$).
\item Every pair of adjacent vertices on path $\pi_i$ is connected by an edge, i.e., $(v_{t}^{i}, v_{t+1}^{i}) \in E, \forall t \in \{0,1,\dots,T^i\}$.
\item $\{\pi_i \mid i\in I\}$ is collision-free.
\end{enumerate}

\section{Related Work}

\subsection{Multi-Agent Path Finding (MAPF)}

MAPF has been proved an NP-hard problem with optimality~\cite{yu2013structure}. It has inspired a wide range of solutions for its related challenges. 
Decoupled strategies, as outlined in \cite{silver2005cooperative,wang2008fast,luna2011push}, approach the problem by independently planning paths for each agent before integrating these paths. 
In contrast, coupled approaches \cite{standley2010finding,standley2011complete} devise a unified plan for all agents simultaneously. There also exist dynamically coupled methods~\cite{sharon2015conflict,wagner2015subdimensional} that consider agents planning independently at first and then together only when needed for resolving agent-agent collisions. 
Among these, Conflict-Based Search (CBS) algorithm \cite{sharon2015conflict} stands out as a centralized and optimal method for MAPF, with several bounded-suboptimal variants such as ECBS~\cite{barer2014suboptimal} and EECBS~\cite{li2021eecbs}. Some suboptimal MAPF algorithms, such as Prioritized Planning (PP)~\cite{erdmann1987multiple,silver2005cooperative}, PBS~\cite{ma2019searching}, LaCAM~\cite{okumura2023lacam} and their variant methods~\cite{chan2023greedy,li2022mapf,okumura2023lacam3} exhibit better scalability and efficiency. However, these search-based algorithms always face the problem of search space dimensionality explosion as the problem size increases, making it difficult to produce a valid solution within a limited time. Learning-based methods can overcome the dimensionality issue by learning from large amounts of data and addressing the trade-off between low-cost paths and scalability.

\subsection{Lifelong MAPF}

Compared to the MAPF problem, Lifelong MAPF (LMAPF) continuously assigns new target locations to agents once they have reached their current targets. In LMAPF, agents do not need to arrive at their targets simultaneously. There are three main approaches to solving LMAPF: solving the problem as a whole~\cite{nguyen2019generalized}, using MAPF methods but replanning all paths at each specified timestep~\cite{li2021lifelong,okumura2022priority}, and replanning only when agents reach their current targets and are assigned new ones~\cite{vcap2015complete,grenouilleau2019multi}. Some algorithms can solve LMAPF in an offline setting where all tasks are known in advance.  include CBSS~\cite{ren23cbss}, which applies Traveling Salesman Problem (TSP) methods to plan task orders. However, these LMAPF methods also face the same scalability problem as MAPF methods.

\subsection{Learning-based MAPF}

Given the huge success of deep learning, many learning-based MAPF methods have been proposed. Compared to search-based algorithms, these methods can usually complete planning in short time and automatically learn heuristic functions. Some of these methods focus on modifying edge weights in the map, such as the congestion model~\cite{yu2023congestion}, which is a data-driven approach that predicts agents’ movement delays and uses these delays as movement costs, or Online GGO~\cite{ZangAAAI25}, which optimizes edge weights for Lifelong MAPF. However, these methods split MAPF planning into multiple stages, which can lead to a larger optimization search space if one considers both edge-weight design and the MAPF solver simultaneously.

Most other methods focus on the solver side, using imitation learning (IL), reinforcement learning (RL), or both. Learning-based solves can make end-to-end decisions using all available information and can be trained on various data types (e.g., MAPF, TAPF, LMAPF). In contrast, search-based methods often require multi-stage decomposition with hand-crafted heuristics. One early learning-based method for MAPF is PRIMAL~\cite{sartoretti2019primal} which is trained by RL and IL. It is a decentralized algorithm that relies on an FOV around an agent to generate the actions of that agent. MAPF-GPT~\cite{andreychuk2024mapf} is a GPT-based model for MAPF problems, trained by IL on a large dataset. Other approaches incorporate communication mechanisms in a decentralized manner, such as GNN~\cite{li2020graph} and MAGAT~\cite{li2021message}, which employ Graph Neural Networks (GNNs) for communication, and SCRIMP~\cite{wang2023scrimp}, which uses a global communication mechanism based on transformers.

However, all existing learning-based solvers focus on the agents themselves, forcing researchers to design features of agents. This makes it challenging, if not impossible, to develop a centralized policy that can handle varying numbers of agents and map sizes. Our method is the first centralized MAPF solver to overcome the challenge of feature design and to integrate edge-weight design ideas~\cite{zhang2024guidance,chen2024traffic} into a neural-network-based solver.

\begin{figure*}[t!]
\centering
\includegraphics[width=\textwidth]{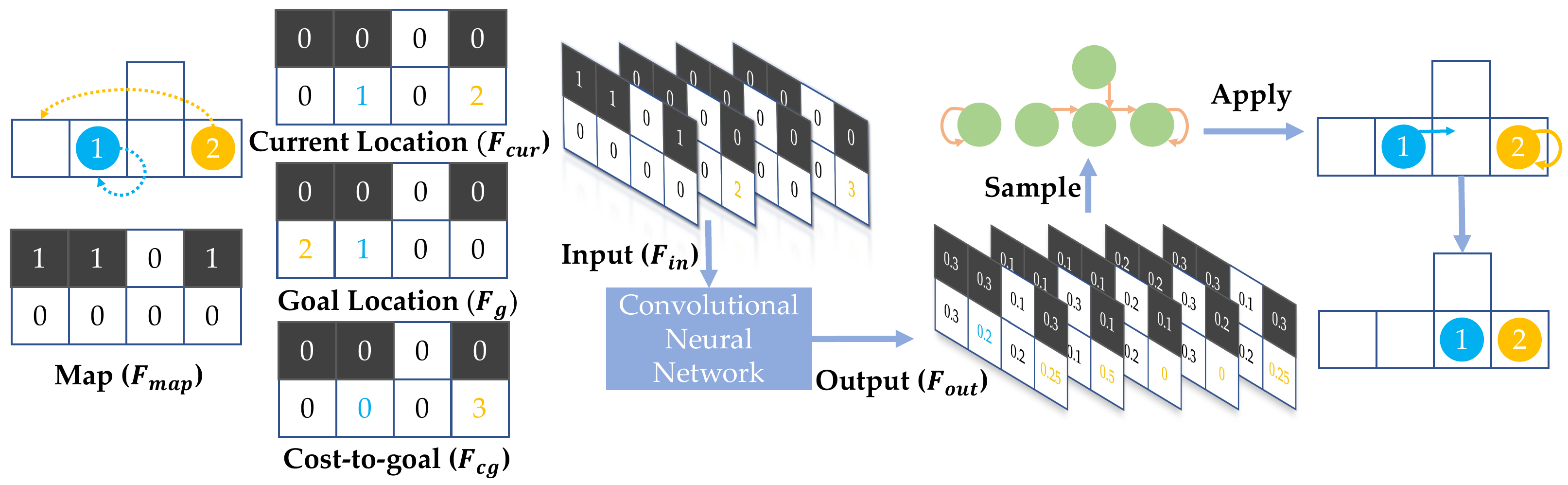} 
\caption{RAILGUN Inference Overview: On the left side (features), there is one current state along with all related input features of size \((n, m, 1)\). These features are then stacked along the last channel to construct the input feature \(F_{in}\) of size \((n, m, k)\). In this example, we have \(n=2\), \(m=4\), and \(k=4\). On the right side (inference), the input feature \(F_{in}\) is fed into a CNN-based neural network, which outputs action probabilities \(F_{out}\) of size \((n, m, 5)\). We sample from \(F_{out}\) to obtain actual actions and then apply the corresponding actions to each agent.}
\vspace{-5mm}
\label{fig:overview}
\end{figure*}

\begin{figure}[htbp]
\centering
\includegraphics[width=0.45\textwidth]{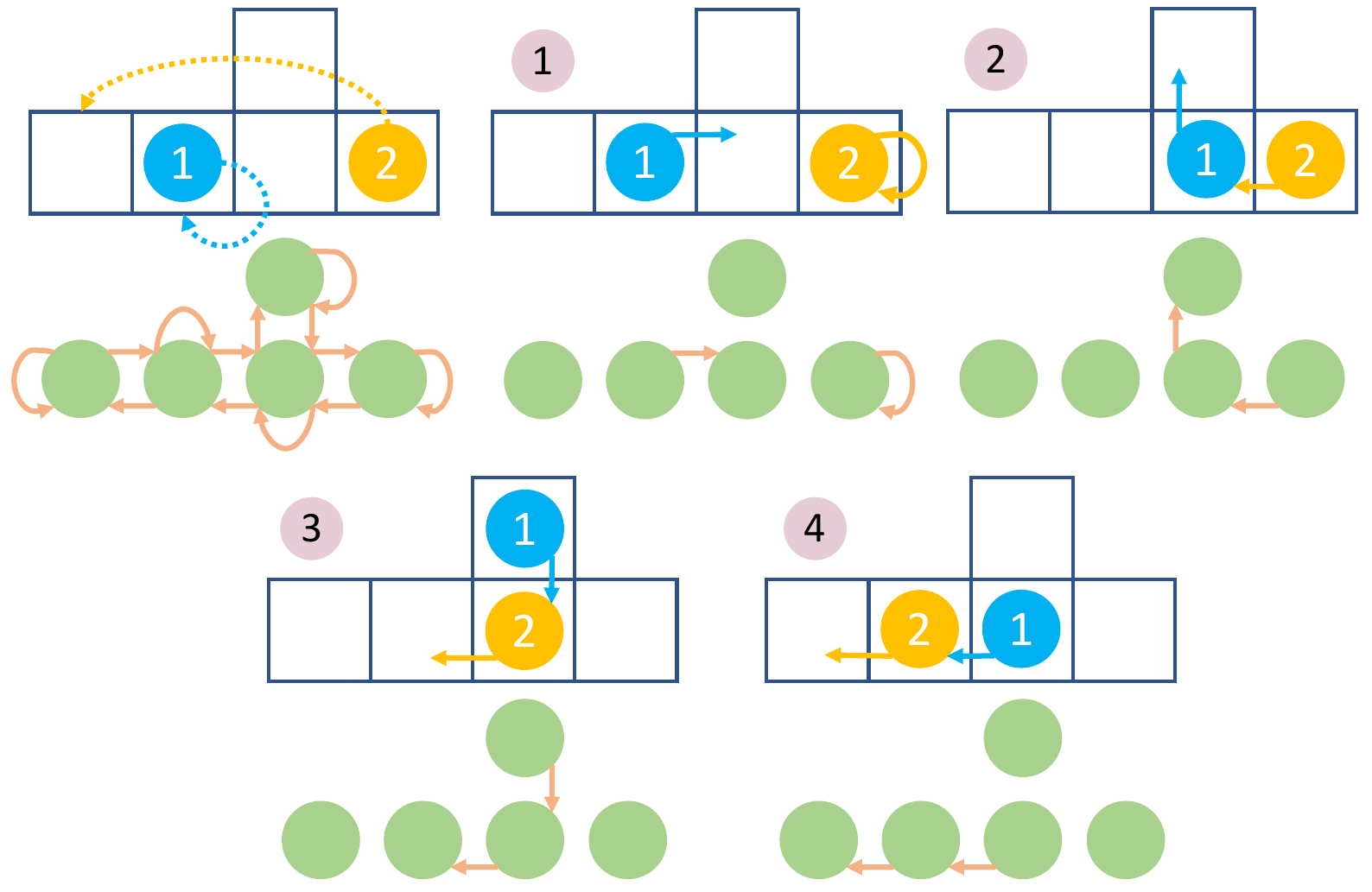} 
\caption{This is an example of how an agent-based solution relates to a series of specialized graphs. The upper-left figure illustrates a testcase we aim to solve, along with a graph where green nodes and orange edges represent map connectivity. The other figures show a valid MAPF solution for this testcase, where agent \(1\) should yield to agent \(2\). At each timestep, each node in the connectivity graph has only one outgoing edge. Here, we draw edges only for nodes occupied by agents, as the outgoing edge for other nodes could be any of the available edges.}
\vspace{-7mm}
\label{fig:trans_plan}
\end{figure}

\section{Method}

\subsection{RAILGUN Overview}

In this section, we introduce our RAILGUN method. First, we discuss why it is difficult to design a learning algorithm for centralized MAPF where policies are agent-based. When focusing on generating actions based on agent features, we need to provide a neural network with at least the agent's start location, goal location, and additional features, amounting to \(k\) scalar variables (\(k \geq 4\)) for one agent. Then the total number of features is at least \(kN\). Consider that the maximum number of agents could be \(N = |V| \approx nm\), where \(n\) and \(m\) are the 2D map dimensions. If we want to handle all possible numbers of agents on a specific map, the total feature size would be \(knm\). This dependence on map size means that we cannot create a policy to cover all different maps if we construct the features agent by agent. That is why there is no centralized learning-based solver and all learning-based MAPF solvers adopt a decentralized approach with a limited FOV for each agent~\cite{sartoretti2019primal,liu2020mapper,li2021message,wang2023scrimp,andreychuk2024mapf}.

Our insight is that in a valid MAPF solution, there will be no collision, which means there can be at most one agent in each map grid cell in each timestep. At any timestep, each agent chooses one of the five edges of its grid cell as its action. Therefore, if we remove all edges that the agents do not use at each timestep, we find that a valid MAPF solution can be viewed as a series of specialized graphs. As shown in \Cref{fig:trans_plan}, these specialized graphs have exactly one edge in every occupied grid cell. Once such a directed graph is given, no MAPF solver is needed, as there is only one possible transition at each timestep. The sequence of these specialized graphs then constitutes a valid MAPF solution.

After converting agent-based solution into a representation as a series of specialized graphs, we use a CNN network to address the challenge of generating these specialized graphs and generalizing across different maps, which we discussed in the previous paragraphs. The input feature is $F_{in}$ with size $(n, m, k)$, and the output feature is $F_{out}$ with size $(n, m, 5)$. Here, $k$ represents the number of feature channels based on the feature design, and $(n, m)$ represents the map size.

As an example shown in \Cref{fig:overview}, to encode an agent’s current location as a feature, we construct a tensor \(F_{cur}\) with size \((n, m, 1)\). In this tensor, \(F_{cur}[i][j] = \text{\textit{idx}}\) if the agent \(\text{\textit{idx}}\) is at position \((i, j)\) in the map; otherwise, \(F_{cur}[i][j] = 0\). Stacking all such feature tensors along the last dimension forms \(F_{in}\) with size \((n, m, k)\).  \(F_{out}[i][j]\) represents the probability distribution over all possible actions at grid cell position \((i, j)\). We use 5 channels because each agent can take one of up to five different actions at each timestep. Thus, if an agent is located at grid cell \((i, j)\), its action probabilities are stored in \(F_{out}[i][j]\). Note that the model only outputs the next action prediction based on the current timestep. Therefore, to obtain a full trajectory, the trained model must be invoked repeatedly.

\subsection{Model Architecture}
In this paper, we use U-Net for RAILGUN, as it is widely used in diffusion models~\cite{ho2020denoising} and includes transposed convolution layers that allows the network to recover the spatial resolution of the input. That’s why RAILGUN can handle input maps of different sizes during inference. We employ the standard U-Net architecture comprising five layers in total. The encoder begins with an initial layer containing 64 channels. At each subsequent layer, the number of channels is doubled while the size of the feature maps is halved. In the decoder, this process is reversed, with the number of channels halved and the spatial resolution doubled at each layer. Notably, bilinear interpolation is not employed in the decoder; instead, we use deconvolution as in original U-Net. At the final layer, the number of channels is reduced to 5, corresponding to the maximum number of possible actions. We should also note that since U-Net uses CNN layers in the encoder, which progressively reduce the spatial dimensions of the feature maps, there is a minimum required input size to ensure valid downsampling operations. For small maps, padding is needed. The resulting RAILGUN model contains approximately 30 million FP32 parameters.

\subsection{Feature Selection}
As shown in \Cref{fig:overview}, we construct the input features from multiple components. We employ five types of features: the map, current locations, goal locations, cost-to-goal, and gradients of cost-to-goal (the last feature is not shown in \Cref{fig:overview} due to space constraints). For the map feature, we use 1 to represent non-traversable grid cells and 0 to represent traversable grid cells. For the current and goal locations, we use the agent's index to indicate which agent occupies a grid cell; otherwise, the grid cell is set to 0. We also attempted encoding agent indices as binary vectors; however, this produces excessively large input features, rendering the model too large to train.

We also use the precomputed shortest path cost as the cost-to-goal feature for each agent which is a widely used feature in learning-based methods, as shown in \Cref{fig:overview}. The gradients of the cost-to-goal, represents the potential direction of next action, are determined by the changes of the cost-to-goal distances. Specifically, we define the changes in cost-to-goal distances from an agent's current cell \((i, j)\) to its adjacent cells as \(\delta_{\text{left}}, \delta_{\text{right}}, \delta_{\text{up}}, \delta_{\text{down}}\). \(\delta < 0\) indicates that the agent is approaching the goal location. The resulting direction, denoted by \(\mathbf{g}_{ij} = \left(\Delta x_{ij}, \Delta y_{ij}\right)\), consists of horizontal and vertical components. For the horizontal component, \(\Delta x_{ij}\) is computed as shown below:
\[
\Delta x_{ij}= \begin{cases}
0 & \text{if } \delta_{\text{left}} \geq 0 \text{ and } \delta_{\text{right}} \geq 0, \\[1ex]
1 & \text{if } \delta_{\text{left}} \geq 0 \text{ and } \delta_{\text{right}} < 0, \\[1ex]
-1 & \text{if } \delta_{\text{left}} < 0 \text{ and } \delta_{\text{right}} \geq 0, \\[1ex]
\text{random}(\pm 1) & \text{if } \delta_{\text{left}} < 0 \text{ and } \delta_{\text{right}} < 0,
\end{cases}
\]
and similarly for the vertical component.

\begin{figure}[tb!]
    \centering
    \begin{subfigure}[b]{0.15\textwidth}
        \centering
        \includegraphics[width=\textwidth]{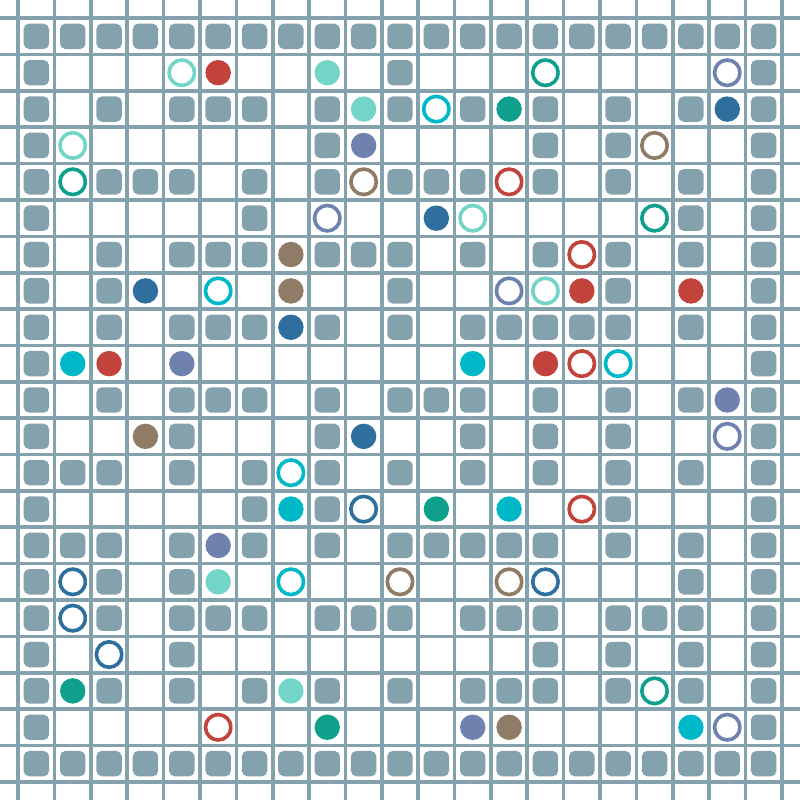}
        \caption{Maze}
        \label{fig:maze}
    \end{subfigure}
    \begin{subfigure}[b]{0.15\textwidth}
        \centering
        \includegraphics[width=\textwidth]{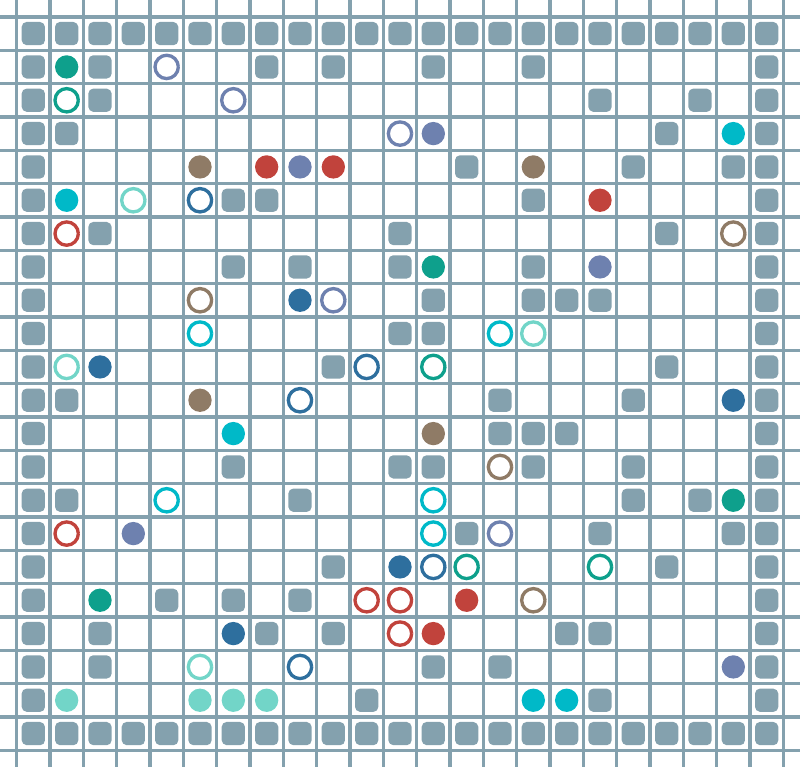}
        \caption{Random}
        \label{fig:random}
    \end{subfigure}
    \begin{subfigure}[b]{0.15\textwidth}
        \centering
        \includegraphics[width=\textwidth]{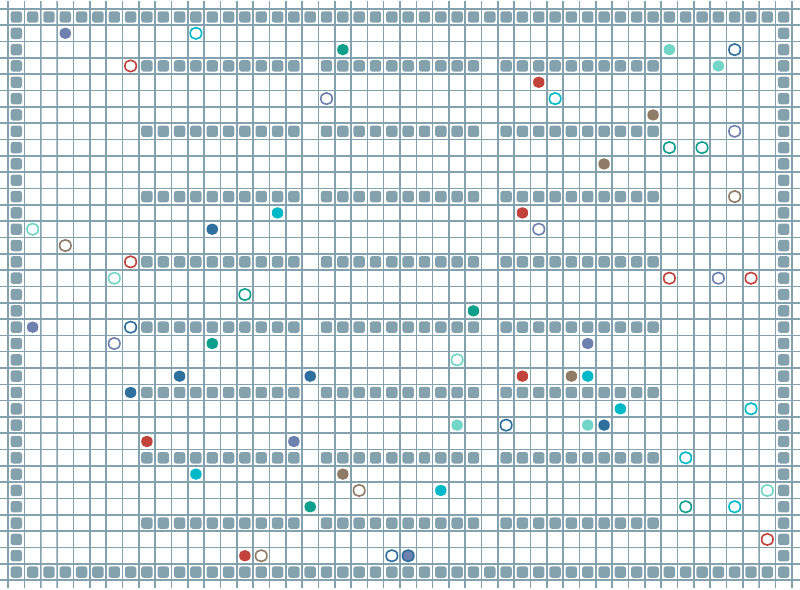}
        \caption{Warehouse}
        \label{fig:warehouse}
    \end{subfigure}
    
    \begin{subfigure}[b]{0.15\textwidth}
        \centering
        \includegraphics[width=\textwidth]{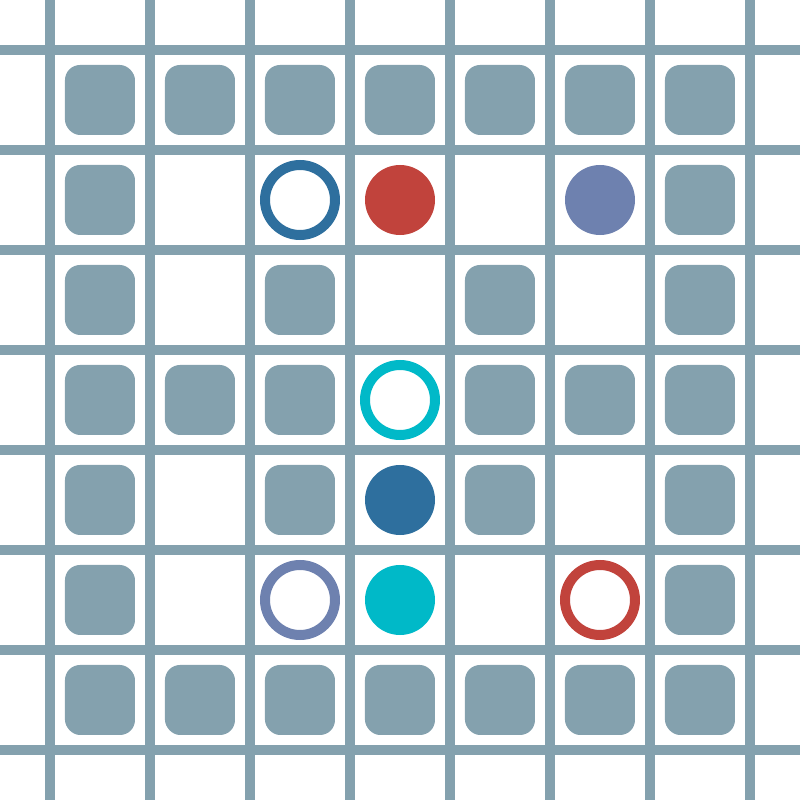}
        \caption{Puzzle}
        \label{fig:puzzle}
    \end{subfigure}
    \begin{subfigure}[b]{0.15\textwidth}
        \centering
        \includegraphics[width=\textwidth]{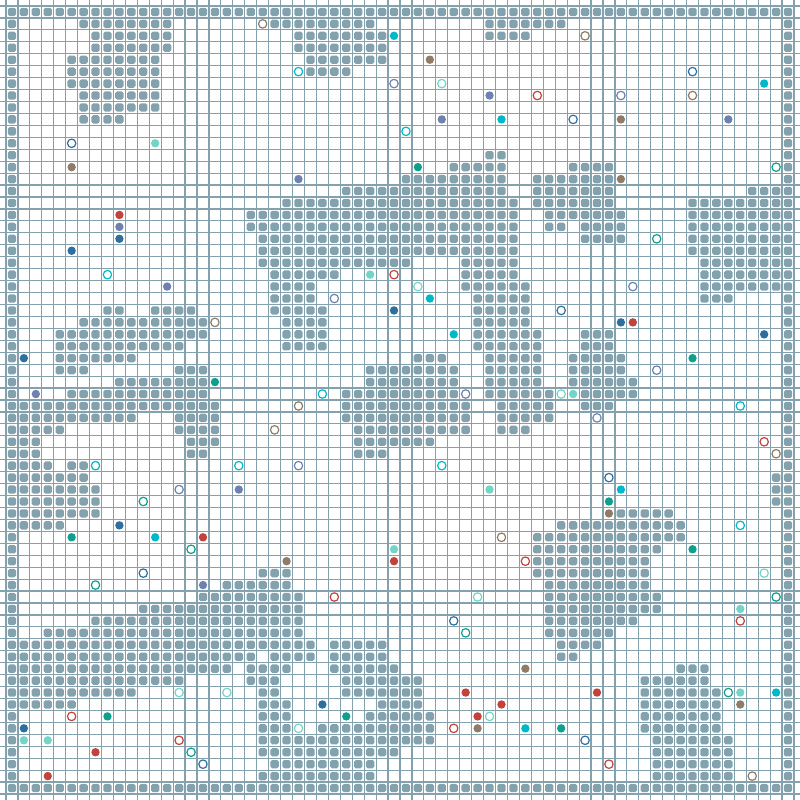}
        \caption{Cities-tiles}
        \label{fig:cities-tiles}
    \end{subfigure}
    \begin{subfigure}[b]{0.15\textwidth}
        \centering
        \includegraphics[width=\textwidth]{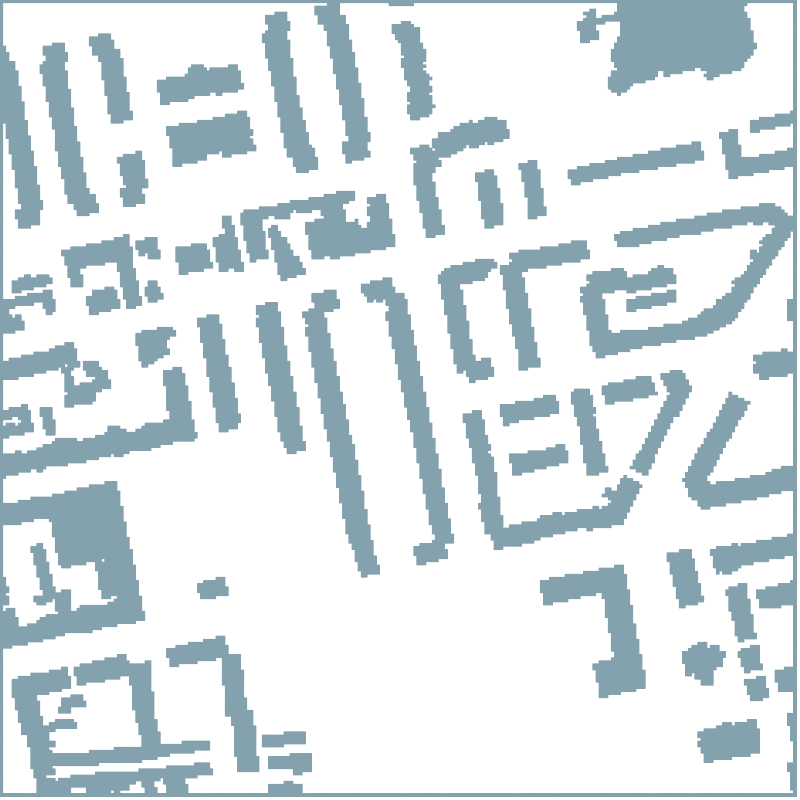}
        \caption{Cities}
        \label{fig:cities}
    \end{subfigure}
    \caption{Examples of POGEMA-tested maps. The six metrics—\textit{Performance}, \textit{Coordination}, \textit{Scalability}, \textit{Cooperation}, \textit{OOD}, and \textit{Pathfinding}—are evaluated on different map sets. Note that Cities-tiles are 64×64 areas selected from larger Cities maps with dimensions of 256×256.}
    \label{fig:maps_examples}
    \vspace{-7mm}
\end{figure}

\section{Experiments \& Results}

\subsection{Training and Testing Settings}
We use the POGEMA~\cite{skrynnik2024pogemabenchmarkplatformcooperative} benchmark to evaluate our method, so for comparison, we only include the methods available in POGEMA. POGEMA includes several different metrics, allowing a fair multi-fold comparison.
For data collection, our training data is primarily generated by LaCAM-v1~\cite{okumura2023lacam}. The model is trained with cross-entropy loss and a batch size of 256. We utilize the AdamW optimizer~\cite{loshchilov2017decoupled} with $\beta$ values set to $(0.9, 0.999)$ and a weight decay of $10^{-3}$. The training process achieves convergence in only six hours, leveraging the power of four NVIDIA A100 GPUs.

For training data, we randomly generate 180 maze maps with $32\times32$ size, each with varying obstacle densities and maze shape. For each map, we randomly generate \{2, 5, 20, 40, 60\} MAPF scenarios with \{16, 32, 64, 96, 128\} agents respectively, for a total of 127 scenarios for each map. We use LACAM-v1~\cite{okumura2023lacam} to compute reference paths for all scenarios as training data, primarily due to its fast data generation speed on large maps.

All experiments\footnote{Our code can be found at Github: https://github.com/TachikakaMin} were conducted on a system running Ubuntu 22.04.1 LTS equipped with an AMD Intel i9-12900K CPU, 128GB RAM and NVIDIA RTX 3080. For the testing phase, the POGEMA benchmark provides a total of 3,376 test cases featuring six different types of maps shown in \Cref{fig:maps_examples}, varying numbers of agents, and different map sizes. POGEMA use six metrics, namely, \textit{Performance}, \textit{Coordination}, \textit{Scalability}, \textit{Cooperation}, \textit{Out-of-Distribution (OOD)} and \textit{Pathfinding}. The relevant equations are as follows:
\begin{gather*}
    \text{Performance} = 
    \begin{cases} 
        SoC_{best}/SoC & \text{if MAPF solved} \\
        0 & \text{if MAPF not solved} \\
        \frac{\text{throughput}}{\text{throughput}_{best}} & \text{if LMAPF}
    \end{cases}\\
\end{gather*}
The \textit{Performance}, \textit{OOD}, and \textit{Cooperation} metrics share the same definitions and primarily evaluate solution quality and success rate across different maps. \(SoC_{best}\) represents the best SoC performance achieved among all tested algorithms.
\begin{gather*}
    \text{Scalability} = \frac{\text{runtime}(\text{agents}_1)/\text{runtime}(\text{agents}_2)}{|\text{agents}_1|/|\text{agents}_2|} \\ 
    \text{Coordination} = 1 - \frac{\text{\# of collisions}}{\lvert \text{agents} \rvert \times \text{episode\_length}}  \\
    \text{Pathfinding} = 
    \begin{cases} 
        \text{SoC}/\text{SoC}_{best} \\
        0 \text{  if path not found}
    \end{cases}
\end{gather*}
\textit{Scalability} is the ratio of algorithm runtimes with different agent numbers with \(|\text{agent}_1| < |\text{agent}_2|\), providing a measure of how the algorithm’s runtime scales as the agent number changes and higher is better. \textit{Coordination} focuses on invalid action frequency produced by learning-based methods. \textit{Pathfinding} indicates the ability of learning-based methods to find the shortest path for a single agent.

\begin{figure}[t!]
    \centering
    \begin{subfigure}[b]{0.4\textwidth}
        \centering
        \includegraphics[width=\textwidth]{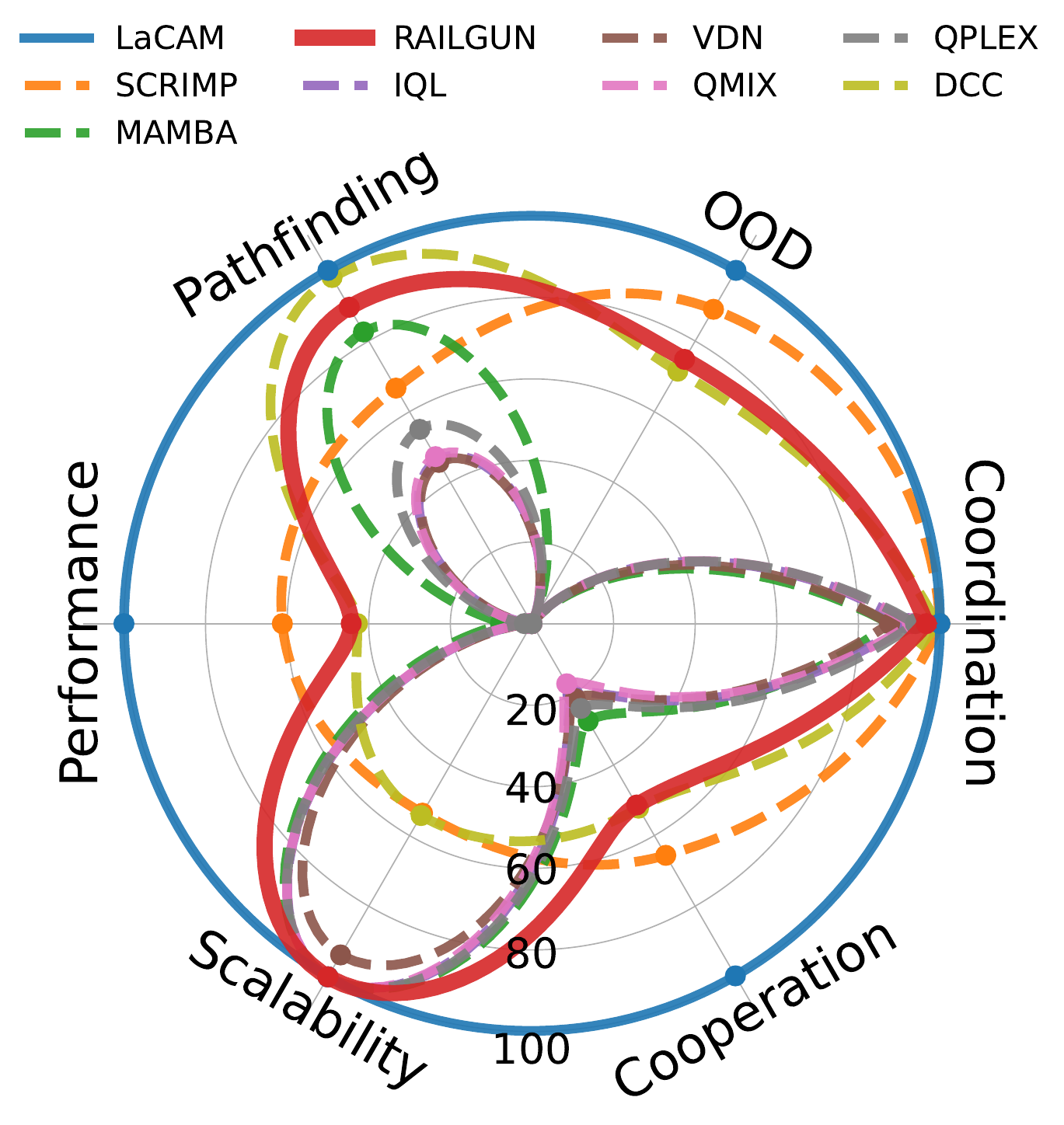}
        \caption{MAPF: RAILGUN outperforms most baseline models in Scalability, Performance, Pathfinding, OOD and Coordination.}
        \label{fig:test_overview_a}
    \end{subfigure}
    
    \vspace{0.1cm}  

    \begin{subfigure}[b]{0.4\textwidth}
        \centering
        \includegraphics[width=\textwidth]{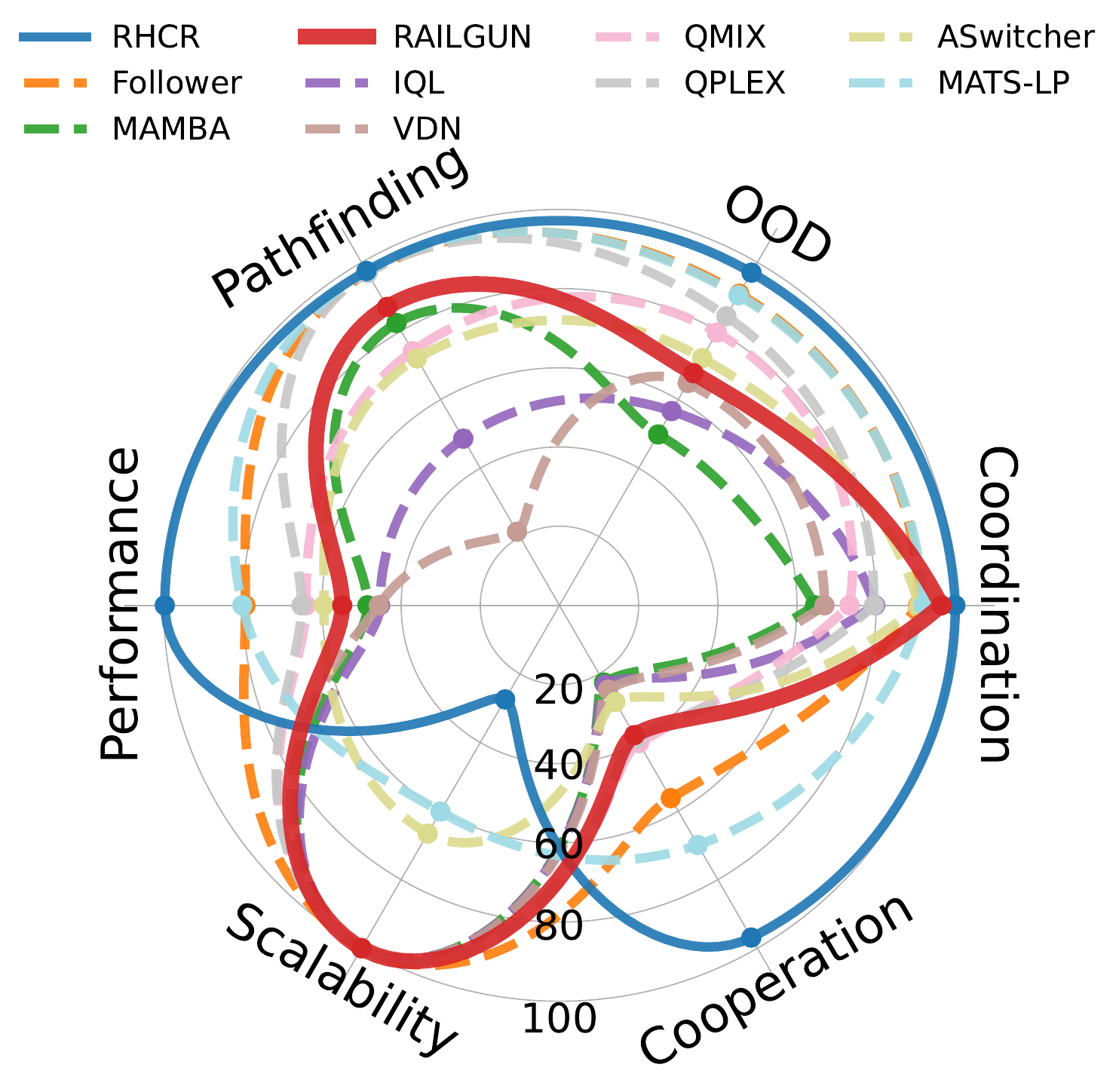}
        \caption{LMAPF: 
RAILGUN still have good zero-shot LMAPF performace in Pathfinding, Coordination, Cooperation and Scalability just training on MAPF dataset.}
        \label{fig:test_overview_b}
    \end{subfigure}

    \caption{POGEMA Test Overview: Performance, OOD, Pathfinding and Cooperation represents solution SoC/throuput quality. Scalability represents runtime respect to agent numbers. Coordination is the probability of invalid actions from learning-based methods.}
    \vspace{-5mm}
    \label{fig:test_overview}
\end{figure}


\subsection{Testing Results}

\Cref{fig:test_overview} presents the performance metrics for RAILGUN and the baseline methods. The learning-based methods include VDN~\cite{sunehag2018value}, QPLEX~\cite{wang2020qplex}, SCRIMP~\cite{wang2023scrimp}, IQL~\cite{tan1993multi}, QMIX~\cite{rashid2020monotonic}, DCC~\cite{ma2021learning}, MAMBA~\cite{egorov2022scalable}, Switcher~\cite{skrynnik2023switch}, Follower~\cite{skrynnik2024learn}, and MATS-LP~\cite{skrynnik2024decentralized}. All these baseline methods are decentralized methods. LaCAM-v3~\cite{okumura2023lacam3} and RHCR~\cite{li2021lifelong} serve as the search-based algorithm baselines in MAPF and LMAPF problems.

In \Cref{fig:test_overview_a}, we observe that RAILGUN achieves high scores across all six metrics. RAILGUN attains the highest score in the Scalability metric because it generates specific directed graphs at each timestep, ensuring that runtime depends only on map size rather than the number of agents in theory. However, even though RAILGUN outperforms or matches the scores of other learning-based methods in most areas, it still exhibits a significant gap with LaCAM in SoC-related metrics. This outcome is expected, as RAILGUN is trained on data generated by LaCAM-v1, and LaCAM-v1 is not designed to achieve the best SoC performance. Mimicking LaCAM-v1 is the top priority of RAILGUN rather than producing a valid solution with the lowest SoC. 

\begin{table*}[htb]
\centering
\setlength{\tabcolsep}{3pt}
\begin{tabular}{c|cccccc|cccccc|cccccc|}
\hline
 & \multicolumn{6}{c|}{CSR} & \multicolumn{6}{c|}{SoC (x1000)} & \multicolumn{6}{c|}{Makespan} \\ 
\hline
\diagbox{Algorithm}{Agents} & 32 & 64 & 96 & 128 & 160 & 192 & 32 & 64 & 96 & 128 & 160 & 192 & 32 & 64 & 96 & 128 & 160 & 192 \\
\hline
LaCAM   & 1.00 & 1.00 & 1.00 & 1.00 & 1.00 & 1.00 & 0.98 & 1.97 & 3.00 & 4.07 & 5.16 & 6.32 & 55.34 & 58.50 & 60.50 & 61.59 & 62.77 & 64.04 \\
SCRIMP  & 1.00 & 1.00 & 1.00 & 0.98 & 0.98 & 0.91 
        & 1.07 & 2.34 & 3.81 & 5.49 & 7.45 & 9.81 
        & 56.54 & 62.48 & 68.17 & 75.25 & 83.43 & 94.73 \\
RAILGUN & 1.00 & 0.97 & 0.73 & 0.13 & 0.01 & - 
        & 1.22 & 2.92 & 5.39 & 9.05 & 13.17 & 17.39 
        & 63.93 & 82.52 & 105.69 & 126.42 & 127.91 & - \\
DCC     & 0.95 & 0.86 & 0.73 & 0.12 & - & - 
        & 1.10 & 2.62 & 4.88 & 7.82 & 11.09 & 14.86 
        & 66.06 & 88.10 & 111.23 & 126.75 & - & - \\
MAMBA   & - & - & - & - & - & - 
        & 2.78 & 7.06 & 11.29 & 15.49 & 19.58 & 23.81 
        & - & - & - & - & - & - \\
IQL     & - & - & - & - & - & - 
        & 4.08 & 8.15 & 12.24 & 16.34 & 20.45 & 24.56 
        & - & - & - & - & - & - \\
VDN     & - & - & - & - & - & - 
        & 3.55 & 7.44 & 11.73 & 16.00 & 20.21 & 24.40 
        & - & - & - & - & - & - \\
QMIX    & - & - & - & - & - & - 
        & 3.67 & 7.56 & 11.64 & 15.85 & 20.06 & 24.28 
        & - & - & - & - & - & - \\
QPLEX   & - & - & - & - & - & - 
        & 3.79 & 7.68 & 11.69 & 15.82 & 20.03 & 24.24 
        & - & - & - & - & - & - \\
\hline
\end{tabular}
\caption{MAPF Scores on Warehouse: Makespan is the latest agent arrival time. ``-'' represents 0 in CSR and 128 in Makespan.}
\vspace{-5mm}
\label{tab:MAPF_performance}
\end{table*}

As shown in \Cref{fig:test_overview_b}, when testing on LMAPF, a completely zero-shot task for RAILGUN, RAILGUN achieves only moderate scores in throughput-related metrics since none of the training data was optimized for throughput. However, this zero-shot test also demonstrates RAILGUN's strong generalization ability across different tasks. RAILGUN also attains high scores in Pathfinding, Coordination, and Scalability. These strengths and weaknesses suggest that, although RAILGUN's overall solution quality remains an issue, it can produce valid solutions in a diverse set of scenarios. Thus, we believe using a dataset optimized for the cost function of interest, combined with applying a task-specific reward function for fine-tuning via RL after the SL process, will help improve the overall solution quality.

\begin{figure}[t!]
    \centering
    \includegraphics[width=0.45\textwidth]{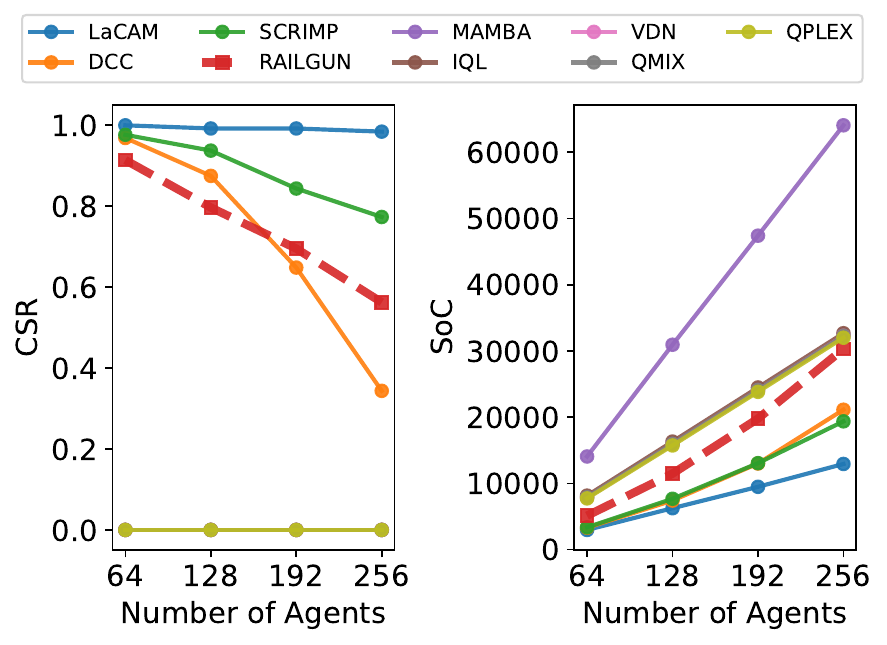}
    \caption{MAPF testing on Cities-tiles: CSR (the success rate at which all agents reach their goal locations; higher is better), SoC (Sum of all agent arrival time; lower is better).}
    \vspace{-5mm}
    \label{fig:MAPF_SOC_CSR}
\end{figure}

\Cref{fig:MAPF_SOC_CSR} presents detailed CSR (see caption) and SoC. Even for unseen maps (Cities-tiles) and larger agent numbers (192 and 256), RAILGUN outperforms other learning-based methods except SCRIMP, achieving up to 60\% CSR. This also shows RAILGUN's strong zero-shot generalization ability in new maps and new agent numbers. Furthermore, we observe that DCC attains a better SoC, despite having a lower CSR. This indicates that in DCC, only a few agents fail to reach their goal locations and the path lengths are shorter than those produced by RAILGUN, highlighting that generating valid solutions is a higher priority for RAILGUN.

\begin{figure}[t!]
    \centering    \includegraphics[width=0.45\textwidth]{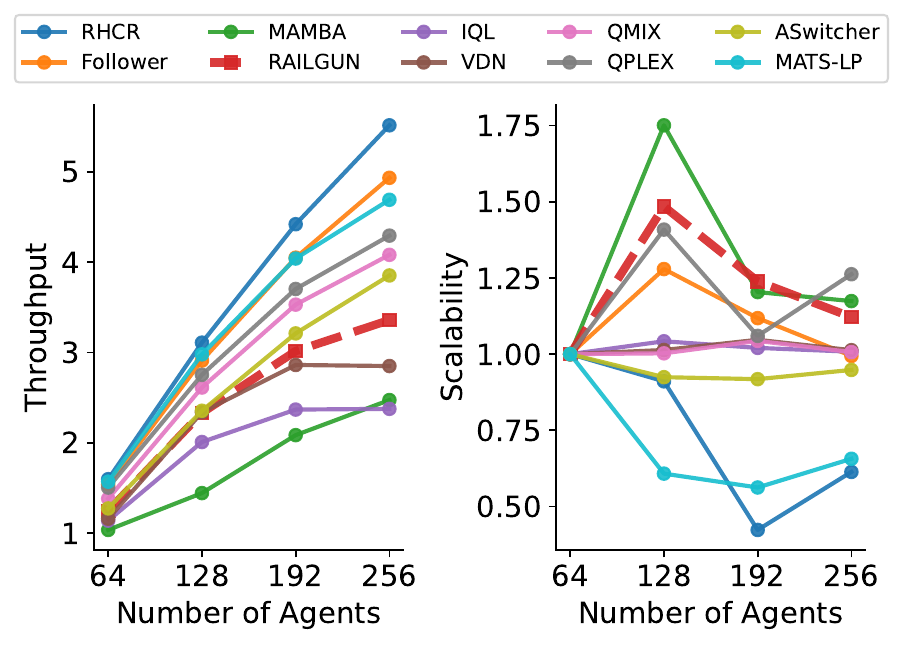}
    \caption{LMAPF Throughput and Scalability Performance on Cities-tiles: Scalability is calculated by previous average per agent runtime divide by current one.}
    \vspace{-5mm}
    \label{fig:LMAPF_Throughput}
\end{figure}

In \Cref{tab:MAPF_performance}, we present the CSR, SoC, and Makespan metrics (see caption) of different algorithms for the Warehouse map. We observe a similar pattern where RAILGUN achieves a higher CSR score but a lower SoC compared to DCC. However, when considering Makespan, RAILGUN outperforms DCC. This indicates that RAILGUN is capable of finding relatively short solutions. As Makespan reflects the latest arrival time, many agents arriving before the last ones contribute to a higher SoC. This may be due to congestion situations, where many agents have a dead lock, and RAILGUN requires additional time to resolve the congestion\footnote{We have a demonstration in our video supplementary material.}. SCRIMP achieves the best MAPF performance among learning-based methods, which may indicate that RAILGUN's supervised learning is not sufficient and that reinforcement learning should also be involved. For LMAPF, as shown in \Cref{fig:LMAPF_Throughput}, RAILGUN's throughput score is better than those of VDN, IQL, and MAMBA. Although it does not achieve the best throughput score overall, its scalability is impressive. \Cref{fig:LMAPF_Throughput} also demonstrates that as the number of agents increases, the average runtime per agent decreases.

\section{Conclusion and Future Work}

In this paper, we propose the first centralized learning-based method, RAILGUN, for the MAPF problem. We found that, rather than predicting actions for individual agents, predicting edge directions for each map grid cell overcomes the difficulties associated with variable input feature dimensions. This finding allows RAILGUN to employ a CNN-based architecture capable of handling maps of any size and any number of agents. In our experiments, RAILGUN demonstrates strong performance across all six metrics in the POGEMA benchmark. Furthermore, its excellent generalization abilities enable it to handle unseen maps, varying agent numbers, and even other tasks such as the LMAPF problem. In future work, we plan to collect higher-quality data to train RAILGUN as a foundation model and apply RL with a task-specific cost function to fine-tune RAILGUN on specific tasks, agent numbers, and map shapes, thereby improving solution quality and success rate in real-world applications.

\section{Acknowledgement}

The research at the University of California, Irvine, the Carnegie Mellon University and the University 
of Southern California was supported by the National Science Foundation 
(NSF) under grant numbers 2328671, 2441629, 2434916, 2321786, 2112533, and 2121028, as 
well as gifts from Amazon Robotics and the Donald Bren Foundation. The views and conclusions contained in this document
are those of the authors and should not be interpreted as representing
the official policies, either expressed or implied, of the sponsoring
organizations, agencies, or the U.S.  government.

\bibliographystyle{IEEEtran} 
\bibliography{strings,myref}

\end{document}